\documentclass[letterpaper,10pt,conference]{ieeeconf}
\IEEEoverridecommandlockouts
\PassOptionsToPackage{svgnames,table}{xcolor}
\usepackage{graphics} %
\usepackage{epsfig} %
\usepackage{mathptmx} %
\usepackage{times} %
\usepackage{amsmath} %
\usepackage{amssymb}  %
\usepackage[utf8]{inputenc} %
\usepackage[T1]{fontenc}    %
\usepackage{url}            %
\usepackage{booktabs}       %
\usepackage{amsfonts}       %
\usepackage{nicefrac}       %
\usepackage{microtype}      %
\usepackage{xcolor}         %
\usepackage{pifont}
\usepackage{tabularx}
\usepackage{array}
\usepackage{graphicx}
\usepackage{float}
\usepackage{wrapfig}
\usepackage{caption, subcaption, overpic, textpos}
\usepackage{makecell}
\usepackage{comment}
\usepackage{multicol,multirow}

\usepackage{xspace}
\usepackage[svgnames,table]{xcolor}
\usepackage{algorithm}
\usepackage{algorithmic}
\definecolor{deemph}{gray}{0.6}
\definecolor{baselinecolor}{gray}{.9}
\definecolor{yellow}{RGB}{218,165,32}
\definecolor{lightcyan}{rgb}{0.88, 1.0, 1.0}
\definecolor{lightskyblue}{rgb}{0.53, 0.81, 0.98}
\definecolor{aliceblue}{rgb}{0.94, 0.97, 1.0}
\definecolor{LightSlateBlue}{RGB}{70,130,180}
\definecolor{DeepBlue}{RGB}{65,100,170}
\definecolor{DeepPurple}{RGB}{136,105,160}
\definecolor{LightGreen}{RGB}{59,125,35}
\definecolor{LightRed}{RGB}{234,66,53}
\definecolor{cvprblue}{rgb}{0.21,0.49,0.74}

\usepackage[nocompress]{cite}

\makeatletter
\let\NAT@parse\undefined
\makeatother
\usepackage[breaklinks,colorlinks,linkcolor=red,urlcolor=magenta,citecolor=cvprblue,bookmarksopen]{hyperref}

\usepackage[capitalize]{cleveref}
\crefname{section}{Sec.}{Secs.}
\Crefname{section}{Section}{Sections}
\Crefname{figure}{Figure}{Figures}
\crefname{figure}{Fig.}{Figs.}
\Crefname{table}{Table}{Tables}
\crefname{table}{Tab.}{Tabs.}

\makeatletter
\DeclareRobustCommand\onedot{\futurelet\@let@token\@onedot}
\def\@onedot{\ifx\@let@token.\else.\null\fi\xspace}

\makeatother

\newcommand{\methodname}{AMS\xspace}

\usepackage{placeins}
\renewcommand{\methodname}{PingpongDuo\xspace}
\usepackage{float}
\newcommand{\na}{\textemdash}

\renewcommand{\dbltopfraction}{0.95}

\begin{document}
\title{\LARGE\bf PingpongDuo: Agile Humanoid Table Tennis
Doubles \\ with Safe Team Coordination }

\author{
\authorblockN{
Haoran Jiang$^{1,2,3}$ \quad Yichao Zhong$^{3}$ \quad Kai Zhang$^{3}$ \quad Cheng Liu$^{3}$ \quad Yixuan Pan$^{3}$ \\
Zhenjie Yang$^{3}$ \quad Jingbo Wang$^{3}$ \quad Tianyu Li$^{2,1}$ \quad Li Chen$^{2
}$
}
\vspace{1mm}
\authorblockA{
$^{1}$Shanghai Innovation Institute \quad $^{2}$Archon Robotics \quad
$^{3}$The University of Hong Kong
}
{\url{https://www.archon.tech/research/PingpongDuo}}
}

\twocolumn[{%
\renewcommand\twocolumn[1][]{#1}%
\maketitle\vspace{-3em}
\begin{center}
\includegraphics[width=0.98\textwidth]{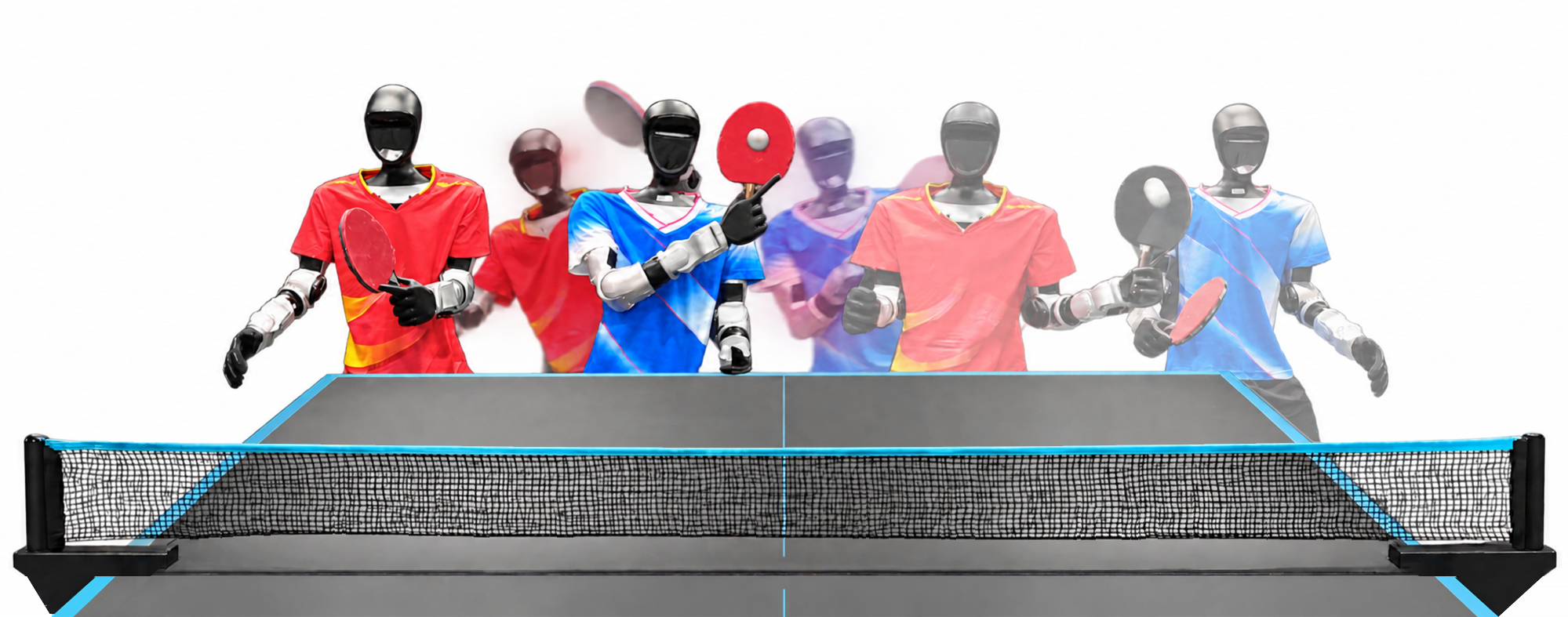}
\captionof{figure}{
\textbf{PingpongDuo: the first
table tennis \textit{doubles} system on Humanoid.}
After returning a ball on the right side of the table, the red robot clears space as the blue robot moves left to receive the next ball, achieving the complete movement
within one second.}
\label{fig:teaser}
\end{center}
}]

\thispagestyle{empty}
\pagestyle{empty}
\flushbottom
\begin{abstract}
Unlike singles, humanoid table tennis doubles requires teammates
to strike in strict alternation while sharing a confined workspace.
Rapid repositioning can compromise striking accuracy, while
uncoordinated teammate movements introduce collision risks.
We present \methodname, a hierarchical framework for agile
humanoid doubles with safe team coordination.
At the team level, a planner jointly selects positioning targets
and uses learned risk estimates to filter risky proposals,
preferentially preserving the current striker's target.
At the individual level, separate footwork and striking skills
combine agile movement with precise racket control.
The skills are guided by human motion references, including
footwork collected from human doubles play, and trained from
states produced by one another to support rapid transitions.
A racket-arm residual further refines strikes after movement
using return feedback from a simulator calibrated with real
ball-flight and contact measurements.
In simulation, PingpongDuo improves the ball-return success rate
by 13.2\%  over the state-of-the-art baseline,
while achieving approximately 2 cm positioning error during clearance.
In real-world experiments, a robot completes a 2.15 m
clearance-and-recovery cycle in 2.4 s with a peak speed of
2.23 m/s, while two robots sustain more than 10 consecutive
strikes in strict alternation.
\end{abstract}

\section{Introduction}
Recent advances in humanoid learning and control~\cite{liao2026beyondmimic, luo2025sonic, pan2025ams} have enabled
robots to perform dynamic interactions with fast-moving
balls~\cite{su2026hitter,zhang2026learning, hu2025pace, ren2026cyboracket, Zhong2026RoboNaldoAS}.
Racket sports combine rapid movement, precise contact
timing, and whole-body balance. Progress in these individual
capabilities provides a foundation for studying humanoid team
sports, where successful execution also depends on coordination
with other players.

Existing humanoid table tennis systems such as
HITTER~\cite{su2026hitter} and SMASH~\cite{ren2026smash} combine
strike planning with learned control to support balance
and consecutive ball returns. Their focus is on individual
striking performance. We study table tennis doubles with two
humanoid teammates on the same side of the table
(\cref{fig:teaser}). Unlike singles, doubles requires teammates
to strike in strict alternation and share the space needed
to approach, strike, and recover.

However, doubles requires each robot to coordinate its movement
with its teammate while remaining ready for its next strike.
 After each strike,
one robot must clear the central hitting area while its teammate moves
into position for the next shot. Insufficient clearance can
obstruct the teammate or cause collisions, whereas excessive
displacement increases the time needed for subsequent recovery.
At the individual level, robots must combine agile footwork with
precise striking. Unlike singles, where movement primarily serves
the incoming shot, doubles robot also requires clearance and recovery
movements that may conflict with the next strike. The robot must
therefore adapt its movement direction and speed while maintaining
accurate racket control.

We present \textbf{\methodname}, a hierarchical framework for agile
humanoid doubles with safe team coordination. A team planner
jointly selects positioning targets, and a learned risk filter
adjusts risky proposals while preferentially preserving the
striker's target. For agile movement and precise striking, we
train separate footwork and striking skills with human motion
references, including footwork collected from human doubles play.
Each skill is trained from states produced by the other to
support repeated transitions. A racket-arm residual further
adjusts the nominal strike to the current robot and ball states
using simulated return feedback. The system
demonstrates more than 10 consecutive alternating strikes and
a 2.15 m clearance-and-recovery cycle completed in 2.4 s in real-world deployment.

In sum, the contributions are:
\begin{itemize}
\item A hierarchical framework for safe team coordination and
agile individual control in humanoid doubles, combining risk-aware
joint positioning with footwork and striking skills trained for
continuous execution.
\item An arm refinement policy that preserves striking
accuracy after rapid footwork by adapting nominal actions to
robot and ball states using simulated return feedback.
\item Simulation and real-world evaluations of individual agility
and team coordination, including the first
real-world demonstration of humanoid-humanoid teammates
completing more than 10 consecutive alternating strikes.
\end{itemize}

\section{Related Work}
\subsection{Humanoid Racket Sports}

Humanoid racket-sport research has advanced whole-body striking,
repositioning, and perception
~\cite{su2026hitter,ren2026smash,ren2026cyboracket,zhang2026learning,huang2026towards}.
Related studies in badminton also investigate coordinated
locomotion and perception
~\cite{Ma2025LearningCB,liu2025badminton,chen2026badminton}.
 Moving beyond individual skills, these capabilities become increasingly important in multi-robot settings, where robots must remain agile while safely coordinating their motions and avoiding collisions with teammates. PingpongDuo further develops a low-level controller that supports agile footwork and striking skills.

Accurate racket-ball contact modeling and closed-loop reinforcement learning (RL) are essential for precise and robust strike performance~\cite{durr2026outplaying}. Human motion references also guide tennis skill learning and
task-specific adaptation~\cite{zhang2026learning,huang2026towards,werner2026tasknpoint}. In the context of table tennis, the small size and high speed of the ball make accurate trajectory modeling particularly important for precise interaction. Ball trajectory and racket--ball interaction modeling support
accurate returns~\cite{durr2026outplaying,nguyen2025spin}. PingpongDuo adopts this trajectory-based formulation. By incorporating simulated return feedback into training, PingpongDuo closes the loop between interaction and striking.

\subsection{Safe Multi-Embodiment Coordination}
Coordination in dynamic sports has been studied in robot
teams~\cite{Li2024MARLadonaT} and human--robot table tennis
doubles~\cite{lee2023collaboration}.
Simulated humanoid football~\cite{Abreu2023DesigningAS} and cooperative quadrupedal
soccer~\cite{su2025toward} combine learned skills with team-level policies.
HCSP~\cite{zhang2025mastering} jointly refines an event-driven team strategy and motor
skills for multi-drone volleyball. Vision-driven humanoid
soccer~\cite{wang2026learning} learns reactive soccer skills while using a high-level
behavior tree to assign roles and coordinate behavior across the team.

Prior works typically deploy robots sparsely in large workspaces, while our table tennis doubles setting requires two humanoids to operate concurrently within a confined workspace. This makes multi-robot co-safety, particularly collision avoidance and motion obstruction, a critical requirement. ABS series ~\cite{He2024AgileBS, zhong2025bridging} integrate agile and recovery policies with reachability estimation to achieve high-speed locomotion, while safe RL approaches~\cite{Yang2022SafeRL} introduce cost functions and recovery policies to enforce safety constraints during learning. In contrast to these single-robot safety formulations, temporal-logic planning~\cite{7497887} and Control Barrier Functions~\cite{CBFmulti,ames2019control} have also been explored for multi-robot coordination and safety. However, acquiring full observations and humanoid Jacobians in the real world is often infeasible. We therefore combine risk estimation with constrained RL~\cite{CPO} for co-safe coordination, while a role-weighted safety filter prioritizes the striker's target and adjusts the non-striker's planned position to maintain ball-striking performance.

\begin{figure*}[t]
\centering
\includegraphics[width=\textwidth]{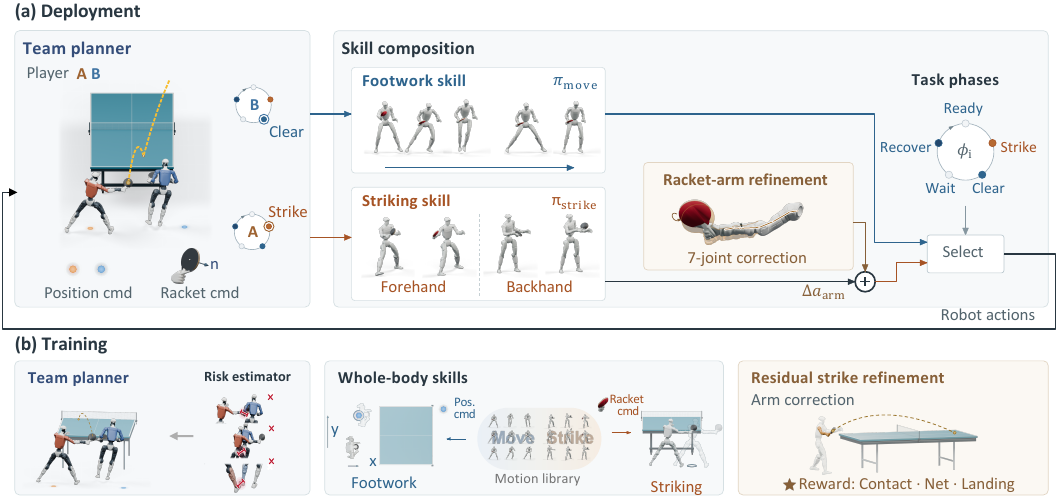}
\caption{\textbf{System overview.}
(a) The team planner coordinates positioning under a learned risk-aware
target filter, while ball prediction and strike planning provide racket
commands (\cref{sec:system_overview}). A state machine selects
footwork or striking actions, with racket-arm refinement applied to the
striking branch.
(b) Training combines safety-aware team coordination (\cref{sec:planner}), skill learning from motion references, and strike refinement using simulated
return feedback (\cref{sec:policy}).}
\label{fig:system}
\end{figure*}

\section{Preliminaries and System Overview}
\label{sec:overview}
We first summarize the ball prediction, strike planning, and striking
policy adopted from prior single-player systems. We then
introduce the doubles task and system architecture.

\subsection{Preliminaries: Foundations for Singles Play}
\label{sec:foundations}
We use a modular formulation that separates ball prediction
and strike planning from low-level motion control
~\cite{su2026hitter,ren2026smash,ren2026cyboracket}.
An adaptive extended Kalman filter estimates ball position and
velocity from successive observations. A physics-based model
predicts the trajectory under gravity, aerodynamic drag, and table
bounces, from which a contact position and time are selected.
Given the predicted incoming velocity and a desired landing location,
a racket--ball interaction model determines the required racket
velocity. For robot $i$, the resulting strike command is
\begin{equation}
c_t^{i,\mathrm{hit}}=
\bigl(p_r^{i,\star},v_r^{i,\star},\tau_t^i\bigr),
\label{eq:strike_command}
\end{equation}
where $p_r^{i,\star}$ and $v_r^{i,\star}$ denote the desired racket
position and velocity at contact, and $\tau_t^i$ is the remaining time to
contact.

The striking policy combines this command with proprioceptive
observations and reference motion information:
\begin{equation}
a_t^{i,\mathrm{hit}}=\pi_{\mathrm{strike}}\bigl(
o_t^{i,\mathrm{prop}},m_t^{i,\mathrm{hit}},c_t^{i,\mathrm{hit}}\bigr),
\label{eq:hit_policy}
\end{equation}
where $o_t^{i,\mathrm{prop}}$ contains proprioceptive observations
and recent control history, $m_t^{i,\mathrm{hit}}$ encodes the
whole-body reference motion, and $a_t^{i,\mathrm{hit}}$ specifies
whole-body joint-position targets.
The reference is selected according to the strike target~\cite{ren2026smash}.
Training combines motion imitation, task rewards, and control
regularization~\cite{liao2026beyondmimic,jiang2025wholebodyvla}: imitation shapes the overall swing, while task rewards
encourage the racket to reach the commanded position and velocity and
to align its face normal with the desired strike direction within a short
window immediately before and after the commanded contact time.

These single-player components do not explicitly coordinate the
non-striker's clearance, so independent positioning can lead to obstruction or
collisions. Doubles also requires frequent repositioning, for which
dense movement rewards can conflict with striking rewards concentrated
around contact.

\subsection{Doubles Task and System Overview}
\label{sec:system_overview}
We consider two humanoid robots $i\in\{A,B\}$ on the same side
of the table. Let $x_t^i$ denote robot $i$'s state,
$b_t$ the ball state, and $\phi_t^i$ its state-machine phase.
$h_k\in\{A,B\}$ identifies the striker for the
$k$th incoming ball. Strict alternation requires $h_{k+1}\ne h_k$.
The other robot is the non-striker. We use $t$ for low-level control
time, $n$ for team-planning updates, and $k$ for incoming-ball events.
After each strike, the roles alternate: the previous striker becomes the
non-striker and clears space for the new striker. Positioning
must therefore balance clearance for the upcoming shot with timely recovery
between shots.

Our framework combines team planning with footwork and striking
skills (\cref{fig:system}). The single-player planning module
provides racket contact targets, while a team planner jointly
selects lateral base targets for both robots. A learned risk estimator supports
a role-weighted target filter that adjusts the paired targets while
preferentially preserving the striker's target. Each skill is trained from states produced by the
other to support repeated transitions. A racket-arm residual uses
robot and ball feedback to refine the strike after movement.

During execution, a state machine combines the accepted positioning
targets with the racket commands from ball prediction and strike
planning. It tracks each robot's phase and enforces the alternating
strike order. For the designated robot, a strike command takes
priority over ongoing clearance, recovery, or waiting and immediately
activates the striking skill. Otherwise, after completing its strike
and follow-through, the robot proceeds through clearance, waiting,
and recovery, then holds its ready position until its next strike
command. The corresponding low-level skill executes each phase.

\section{Safety-Aware Team Coordination}
\label{sec:planner}
Between incoming balls, fixed rules guide the next striker toward
the table center and its teammate outward for clearance.
The RL planner is invoked per ball, using the
predicted strike command to select both robots' positioning
targets. These targets are checked by the risk filter before
execution.

\subsection{Team Positioning Policy}
At the planning update triggered by each incoming ball,
the team positioning policy proposes
\begin{equation}
u_n=\bigl(y_n^A,y_n^B\bigr)^\top
\sim\pi_{\theta}(\,\cdot\mid o_n^{\mathrm{plan}}),
\label{eq:planner_action}
\end{equation}
where $u_n$ is the paired lateral base target and $\theta$ denotes the policy
parameters. Each robot is constrained to remain within a longitudinal (x) range
suitable for striking, while lateral targets $y_n^i$ guide
teammate clearance and recovery for subsequent shots. The planner observation $o_n^{\mathrm{plan}}$ contains the
strike command computed from the latest ball observations
using the physics-based prediction and planning module
(\cref{eq:strike_command}), together with both robots'
current base positions and the previously accepted target pair. The striker's component supports the upcoming strike, whereas
the non-striker's component provides clearance and supports subsequent recovery.

We train $\pi_\theta$ over sequences of incoming balls using constrained
policy optimization based on PPO~\cite{schulman2017proximal,CPO}, while
holding the low-level controller and risk estimator fixed. The objective
maximizes the accumulated team task return subject to a limit on the expected
cost of inter-robot separation violations. At deployment, each proposal $u_n$
is evaluated by the risk-aware target filter before execution.

\subsection{Learned Risk Estimation}
Let $\mathcal Q=\{\mathrm{hand},\mathrm{foot},\mathrm{base}\}$ index the
monitored inter-robot separations. For a candidate target pair executed over
rollout interval $\mathcal I_n$, we assign the binary labels
\begin{equation}
z_{n,q}=\mathbb{I}\!\left[
\min_{t\in\mathcal I_n}\delta_q(t)<\delta_q^{\min}\right],
\qquad q\in\mathcal Q,
\label{eq:risk_labels}
\end{equation}
where $\delta_q(t)$ is the corresponding separation and
$\delta_q^{\min}$ is its violation threshold. The dataset
$\mathcal D_{\mathrm{risk}}=\{(o_n^{\mathrm{risk}},u_n,z_n)\}$ is collected
from closed-loop simulated rollouts with the low-level controller fixed and
targets generated by random and perturbed heuristic planners. An MLP
$r_\psi(o_n^{\mathrm{risk}},u_n)$ is trained in a supervised manner to predict
the per-separation risk vector. The scalar score used by the target filter is
the mean
$\bar r_\psi(o,u)=|\mathcal Q|^{-1}\sum_{q\in\mathcal Q}r_{\psi,q}(o,u)$.

\subsection{Role-Weighted Target Filtering}
If $\bar r_\psi(o_n^{\mathrm{risk}},u_n)\leq\eta$, the proposal is accepted
without modification. Otherwise, the filter searches for the closest
admissible target pair whose predicted mean risk satisfies the threshold:
\begin{equation}
\begin{aligned}
\tilde u_n=\arg\min_{u\in\mathcal U}\quad&
w_{\mathrm{str}}\bigl(u_{\mathrm{str}}-u_{n,\mathrm{str}}\bigr)^2
+w_{\mathrm{nstr}}\bigl(u_{\mathrm{nstr}}-u_{n,\mathrm{nstr}}\bigr)^2\\
\text{subject to}\quad&
\bar r_{\psi}(o_n^{\mathrm{risk}},u)\leq\eta.
\end{aligned}
\label{eq:projection}
\end{equation}
Here $\mathcal U$ specifies the admissible lateral-target bounds, while
$\mathrm{str}$ and $\mathrm{nstr}$ denote the striker and non-striker
components. Choosing $w_{\mathrm{str}}>w_{\mathrm{nstr}}>0$ assigns a larger
modification cost to the striker's target because changing its position near
contact can disrupt the immediate return. The role assignment follows $h_k$
and changes after each incoming ball. We evaluate candidate target pairs by
grid search over $\mathcal U$. If no admissible pair is found, both robots
return to their predefined default targets and the next strike handoff is
withheld.

\section{Learning Agile Footwork and Precise Striking}
\label{sec:policy}
Doubles requires rapid repositioning between successive strikes while
maintaining accurate racket control. We train footwork and striking
skills separately, incorporate states from the preceding skill into
their training, and refine the nominal strike using simulated
ball-interaction outcomes.

\subsection{Footwork and Striking}
We collect whole-body motion data from human table tennis doubles
and segment the recordings into around 200 footwork clips covering different
movement distances. The clips capture movements for
strike preparation, non-striker clearance, and recovery. We use GMR~\cite{araujo2025gmr} to retarget
them to the robot and annotate each clip with its lateral displacement.
Retaining the temporal structure of these motions provides references
for completing the required movements within the time available
between strikes.
These references guide coordinated stepping and body motion
across different repositioning distances, providing motion
priors for both short adjustments and larger lateral movements.

At the start of each movement segment, we select a reference according
to the commanded displacement:
\begin{equation}
\begin{aligned}
m^\star&=\arg\min_{m\in\mathcal M(\Delta y_t^i)}
\left|\Delta y_t^i-\Delta y_m\right|,\\
\Delta y_t^i&=y_{t,\mathrm{cmd}}^i-y_t^i,
\label{eq:footwork_matching}
\end{aligned}
\end{equation}
where $y_{t,\mathrm{cmd}}^i$ and $y_t^i$ are the commanded and current
lateral base positions, $\Delta y_m$ is the displacement of motion clip
$m$, and $\mathcal M(\Delta y_t^i)$ contains clips for the requested
direction.
The team planner determines where the robot should move, while
motion matching selects a reference for executing.

The footwork policy $\pi_{\mathrm{move}}$ receives proprioceptive observations,
a movement command, and the matched reference. Its training combines
position and velocity tracking, motion imitation, stability objectives,
and control regularization. Movement phases emphasize reaching the
target, while waiting phases emphasize braking and maintaining a
ready posture. Target distances and waiting durations are varied
to cover the movement and holding conditions in doubles.

We combine footwork with the striking skill introduced in
\cref{sec:foundations}:
\begin{equation}
a_t^{i,\mathrm{nom}}=
\begin{cases}
\pi_{\mathrm{strike}}\bigl(o_t^{i,\mathrm{prop}},
m_t^{i,\mathrm{hit}},c_t^{i,\mathrm{hit}}\bigr),
& \phi_t^i\in\Phi_{\mathrm{hit}},\\
\pi_{\mathrm{move}}\bigl(o_t^{i,\mathrm{prop}},
m_t^{i,\mathrm{move}},c_t^{i,\mathrm{move}}\bigr),
& \text{otherwise}.
\end{cases}
\label{eq:skills}
\end{equation}
Here the roman superscripts $\mathrm{hit}$ and $\mathrm{move}$ identify
the skill, while $i$ identifies the robot. The variables
$m_t^{i,\mathrm{hit}}$ and $m_t^{i,\mathrm{move}}$ denote the respective
reference-motion features, and $c_t^{i,\mathrm{move}}$ specifies the movement
command. The phase set $\Phi_{\mathrm{hit}}$ includes striking and a brief
follow-through.
Footwork handles preparation, clearance, recovery, and waiting.
Both skills produce whole-body joint actions.

We train the skills in a canonical configuration with right-handed
striking and leftward clearance. A mirrored configuration uses
left-handed striking and rightward clearance for the robot on the
opposite lateral side of the same table half. Mirroring is applied
consistently to observations, commands, reference motions, and joint
actions, allowing both robots to share the trained policies.

\begin{figure*}[t]
    \centering
    \includegraphics[width=\textwidth]{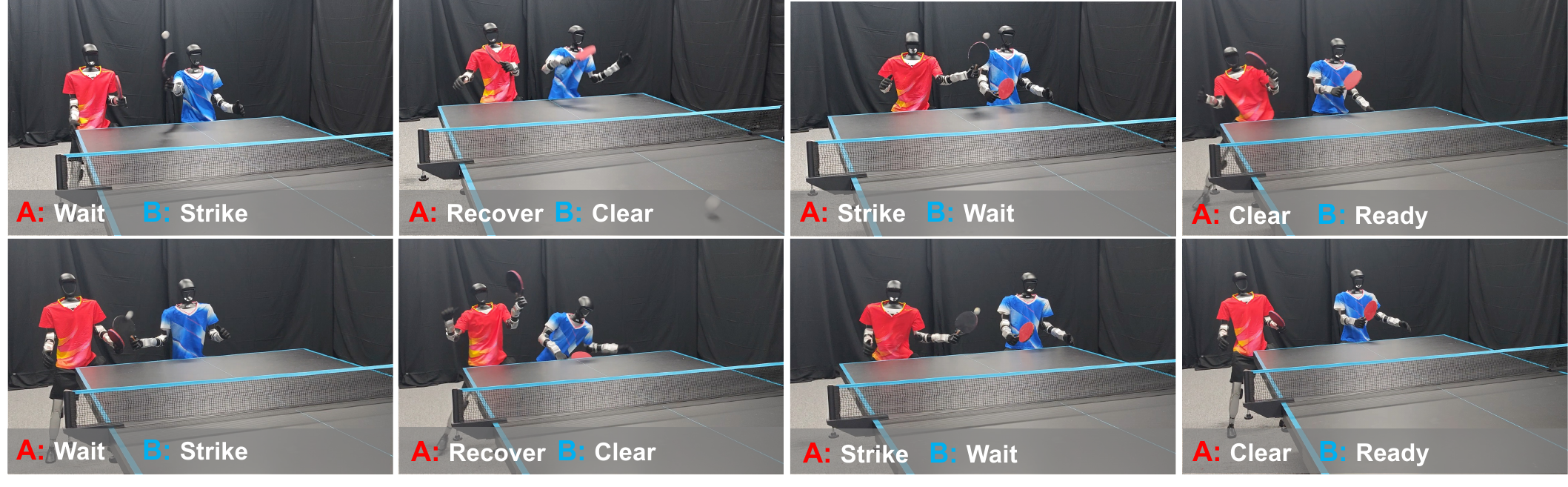}
    \caption{\textbf{Real-world humanoid doubles.} Four consecutive alternating returns
in a rally with human. Each striker clears space as its teammate prepares
for the next shot.}
    \label{fig:multi_robot_pingpong}
\end{figure*}

\subsection{Training for Skill Transitions}
Directly composing independently trained skills can expose each
policy to body states outside its training distribution. Footwork
must begin from the posture and velocity left by a swing, while
striking may begin before the robot has fully settled after movement.
We therefore use states produced by each skill to initialize
training of the other:
\begin{equation}
x_0^{\mathrm{move}}\sim
\mathcal D_{\mathrm{hit}\rightarrow\mathrm{move}},
\qquad
x_0^{\mathrm{hit}}\sim
\mathcal D_{\mathrm{move}\rightarrow\mathrm{hit}},
\label{eq:transition_states}
\end{equation}
where $\mathcal D_{\mathrm{hit}\rightarrow\mathrm{move}}$ contains states
collected after striking and follow-through, and
$\mathcal D_{\mathrm{move}\rightarrow\mathrm{hit}}$ contains states reached
during footwork.

The footwork policy learns to initiate clearance and recovery from
post-strike states. Conversely, the striking policy learns to execute
its racket command from movement-generated states. These initial
states include body position, orientation, linear and angular
velocity, and joint positions and velocities, preserving the dynamics
inherited from the preceding motion.

During execution, control histories are maintained across skill
switches, and reference motions are transitioned over a short
interval. A strike can begin during repositioning, and footwork
resumes after follow-through. This supports repeated movement,
striking, and clearance without resetting the robot to a standard
pose between phases.

\subsection{Strike Refinement with a Racket-Arm Residual}
Reference motions are matched to the strike target, but their
execution also depends on the robot's current body state.
Residual body velocity or an unsettled posture after footwork
can make the selected motion less suited to the current conditions.
Furthermore, position, velocity, and orientation errors jointly
affect the realized contact and outgoing ball trajectory, so
smaller racket-tracking errors do not necessarily yield a better
return. We therefore use ball-interaction feedback to refine
the nominal strike. With the nominal skills fixed, we learn
a bounded correction over the seven racket-arm joints, following~\cite{durr2026outplaying, zhang2026learning}:
\begin{equation}
\begin{aligned}
\Delta a_{t,\mathrm{arm}}^i
&=\pi_{\mathrm{res}}\bigl(o_t^{i,\mathrm{hit}},
a_t^{i,\mathrm{nom}},\hat b_t^i\bigr),\\
a_t^{i,\mathrm{exec}}
&=a_t^{i,\mathrm{nom}}
+\alpha g_t^iS_{\mathrm{arm}}^i\Delta a_{t,\mathrm{arm}}^i.
\end{aligned}
\label{eq:residual}
\end{equation}
The context $o_t^{i,\mathrm{hit}}$ contains proprioceptive observations,
reference motion information, and strike commands; $\hat b_t^i$
contains robot $i$'s estimate of ball position and velocity. The residual
$\Delta a_{t,\mathrm{arm}}^i$ is a bounded seven-dimensional correction for
the racket arm. The selector $S_{\mathrm{arm}}^i$ embeds it into the whole-body
action, and $\alpha$ controls the overall correction magnitude.
A phase-dependent temporal gate $g_t^i$ gradually activates and
removes the correction around contact.

\begin{figure}[!t]
    \centering
    \includegraphics[width=0.48\textwidth]{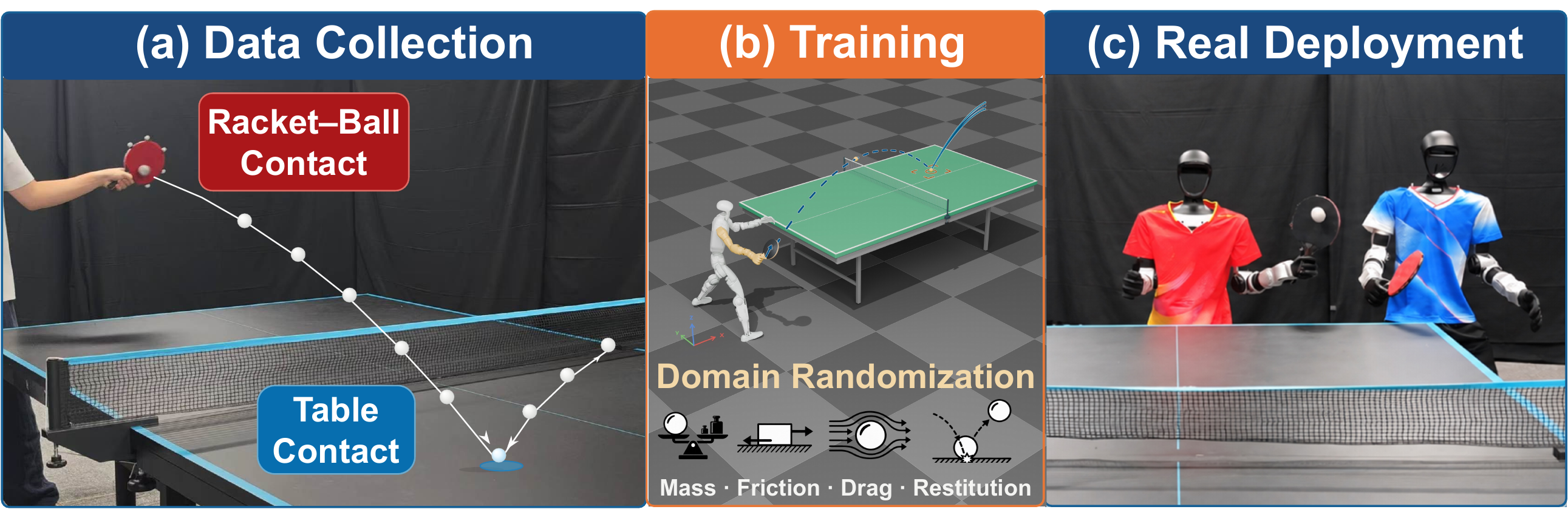}
    \caption{
    \textbf{Real-to-sim-to-real training pipeline for strike refinement.}
    (a) We collect motion-capture trajectories covering racket--ball
    and table contacts to identify free-flight and contact dynamics.
    (b) The identified dynamics define the nominal simulator, where the
    refinement policy is trained with domain randomization over ball mass,
    friction, air drag, and restitution.
    (c) The trained refinement policy is transferred to the real robots
    without further adaptation.
    }
    \label{fig:real2sim2real}
\end{figure}

\textbf{Real-to-sim-to-real.}
As shown in \cref{fig:real2sim2real}, we calibrate the
ball-interaction simulator using staged motion-capture experiments.
We first identify the effective aerodynamic drag coefficient,
table restitution, and friction parameters from vertical drops
and oblique ball--table impacts. Grid search selects parameters
that minimize a weighted discrepancy between simulated and
measured motion. The fitting objectives include time-aligned
trajectory errors, rebound-height errors, and contact-time errors.
For oblique impacts, post-contact velocity errors further constrain
the tangential response, which vertical drops alone cannot identify.

With flight and table dynamics fixed, we fit the racket-contact
model using moving-racket experiments. The fitted parameters
describe speed-dependent normal restitution and effective
tangential attenuation. Incoming and outgoing ball velocities
are expressed relative to the moving racket contact point.
We use grid search followed by local grid refinement to minimize
a weighted sum of outgoing normal-velocity, tangential-speed,
and direction errors. This staged procedure reduces compensation
between flight and contact-model errors. The fitted parameters
define the nominal simulator; separate dynamics randomization
covers modeling error and physical variation during policy training.

\begin{table*}[t]
\centering
\caption{\textbf{Team-planner comparison in simulation.} Mean $\pm$ SD over
five evaluation seeds.}
\label{tab:planner}
\footnotesize
\setlength{\tabcolsep}{2pt}
\renewcommand{\arraystretch}{1.18}
\renewcommand{\tabularxcolumn}[1]{m{#1}}
\begin{tabularx}{\textwidth}{@{}m{0.25\textwidth}*{7}{>{\centering\arraybackslash\leavevmode}X}@{}}
\toprule
Method & \makecell{Mean returns $\uparrow$}
& \makecell{$S_2$\\(\%) $\uparrow$} & \makecell{$S_3$\\(\%) $\uparrow$}
& \makecell{$S_4$\\(\%) $\uparrow$} & \makecell{$E_p$\\(cm) $\downarrow$}
& \makecell{$E_v$\\(m/s) $\downarrow$} & \makecell{$E_n$\\($^\circ$) $\downarrow$}\\
\midrule
Rule-based & $3.60\!\pm\!{\scriptstyle 0.40}$ & $57.40\!\pm\!{\scriptstyle 4.93}$ & $48.00\!\pm\!{\scriptstyle 5.48}$ & $38.40\!\pm\!{\scriptstyle 4.51}$ & $7.92\!\pm\!{\scriptstyle 0.21}$ & $0.56\!\pm\!{\scriptstyle 0.04}$ & $14.34\!\pm\!{\scriptstyle 2.22}$\\
Rule-based + Heuristic CBF & $3.79\!\pm\!{\scriptstyle 0.38}$ & $57.40\!\pm\!{\scriptstyle 4.93}$ & $48.00\!\pm\!{\scriptstyle 5.48}$ & $39.20\!\pm\!{\scriptstyle 5.36}$ & $7.96\!\pm\!{\scriptstyle 0.23}$ & $0.56\!\pm\!{\scriptstyle 0.04}$ & $14.21\!\pm\!{\scriptstyle 2.17}$\\
Rule-based + Learned Risk & $3.79\!\pm\!{\scriptstyle 0.36}$ & $57.40\!\pm\!{\scriptstyle 4.93}$ & $48.00\!\pm\!{\scriptstyle 5.48}$ & $38.80\!\pm\!{\scriptstyle 3.70}$ & $7.94\!\pm\!{\scriptstyle 0.23}$ & $0.56\!\pm\!{\scriptstyle 0.04}$ & $\mathbf{14.19}\!\pm\!{\scriptstyle 2.48}$\\
\rowcolor{black!5}
\makecell[l]{\textbf{RL + Learned Risk} \textbf{(ours)}} & $\mathbf{4.87}\!\pm\!{\scriptstyle 0.74}$ & $\mathbf{64.00}\!\pm\!{\scriptstyle 4.30}$ & $\mathbf{54.80}\!\pm\!{\scriptstyle 4.82}$ & $\mathbf{46.60}\!\pm\!{\scriptstyle 5.68}$ & $\mathbf{7.42}\!\pm\!{\scriptstyle 0.09}$ & $\mathbf{0.54}\!\pm\!{\scriptstyle 0.05}$ & $14.72\!\pm\!{\scriptstyle 2.33}$\\
\bottomrule
\end{tabularx}
\end{table*}

\begin{figure*}[t]
\centering
\includegraphics[width=\textwidth]{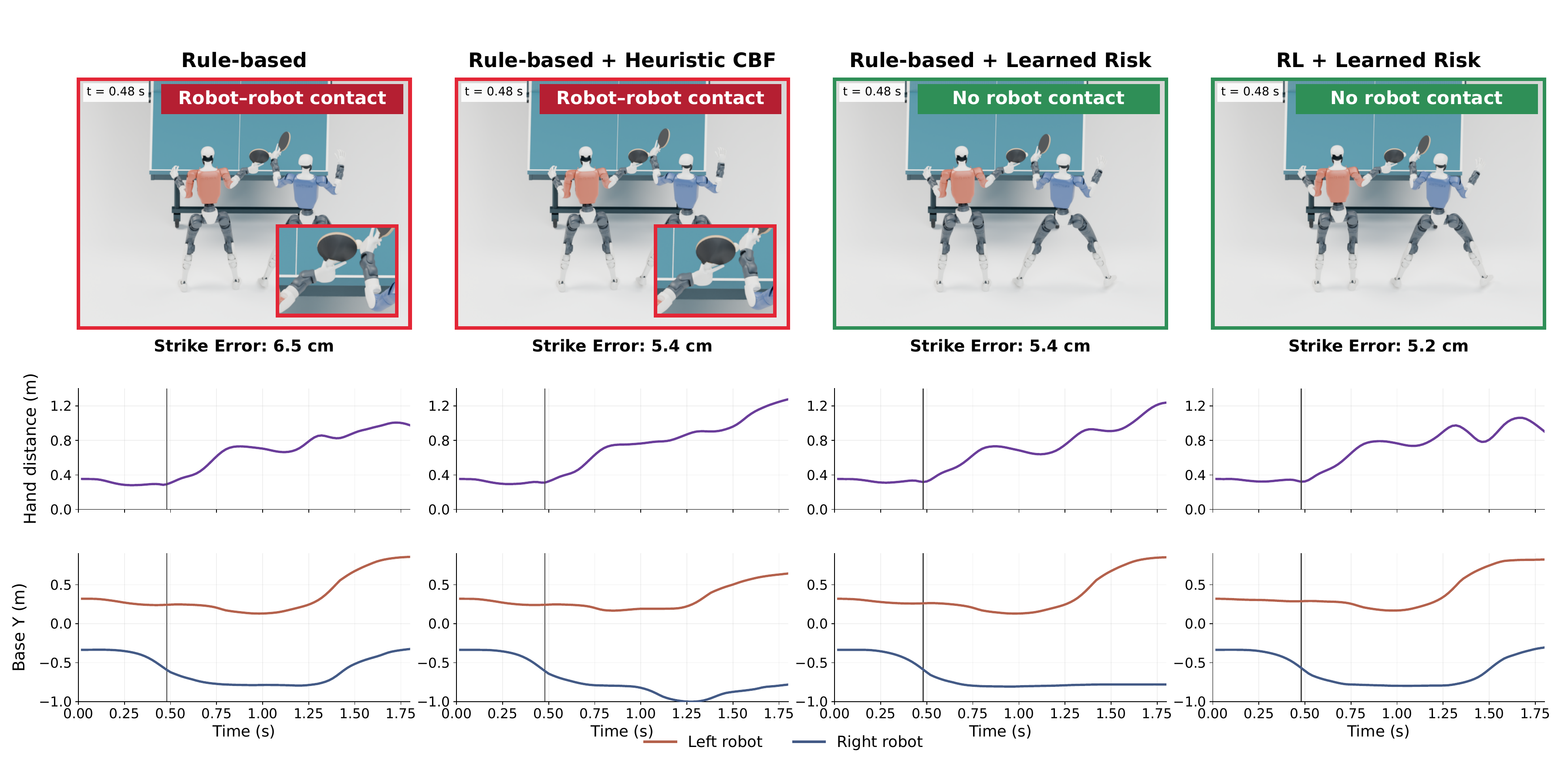}
\caption{\textbf{Planner comparison with baselines under the same coming balls and robot initialization.} Top: simulated execution snapshot at the collision moment of rule-based and rule-based+heuristic CBF; Mid: inner hand distance timeline curve; Down: two robots' lateral base $y$-position timeline curve. Red boxes mark robot collision.}
\label{fig:team}
\end{figure*}

We train the refinement policy with PPO~\cite{schulman2017proximal} on strikes following
footwork, using varied incoming balls and target landing locations.
This exposes the policy to residual body motion after footwork,
allowing it to adapt the strike to the robot's state at contact.
For each ball trial, we randomize ball mass, aerodynamic drag,
table restitution, and racket-contact responses around the nominal
simulation. The corrected arm motion determines the racket state at
impact, while simulated racket contact, net crossing, and landing
provide task-level feedback. Rewards favor valid returns and, for
successful returns, landing accuracy and net-clearance margin.
Misses, low net crossings, out-of-bounds returns, and unintended
ball--body contacts are penalized. Auxiliary racket-tracking and
stability terms are retained together with regularization on the
correction magnitude. Ball-observation perturbations and actuator
delays provide additional variation during training. The resulting
refinement policy is then deployed on the real robot without further
adaptation.

\makeatletter
\setlength{\@dblfptop}{0pt}
\setlength{\@dblfpsep}{8pt}
\setlength{\@dblfpbot}{0pt plus 1fil}
\makeatother

\setcounter{dbltopnumber}{2}
\renewcommand{\dbltopfraction}{0.95}

\section{Experiments}
\label{sec:experiments}
We evaluate \methodname through three main questions:
(1) Can joint positioning and risk filtering support safe,
continuous alternating play?
(2) Does skill composition combine agile footwork with precise
striking more effectively than a joint policy?
(3) Does arm refinement improve ball-return success when
striking from moving states?

\begin{table}[!t]
\centering
\caption{\textbf{Evaluation metrics.}}
\label{tab:metrics}
\footnotesize
\setlength{\tabcolsep}{3pt}
\renewcommand{\arraystretch}{1.08}
\begin{tabularx}{\linewidth}{@{}lcX@{}}
\toprule
Metric & Unit & Definition\\
\midrule
Mean returns & -- & Mean consecutive legal returns\\
$S_k$ & \% & Percentage of attempts with at least $k$ alternating legal returns\\
$SR_C$ & \% & Percentage of balls contacted by the racket\\
$SR_H$ & \% & Percentage of balls legally returned\\
$E_p$ & cm & $100\|p_r(t^\star)-p_r^\star\|_2$\\
$E_v$ & m/s & $\|v_r(t^\star)-v_r^\star\|_2$\\
$E_n$ & deg & $\frac{180}{\pi}\arccos(|n_r(t^\star)^\top\hat v_r^\star|)$\\
$E_M$ & cm & Lateral target error at the end of clearance\\
MPJPE & mm & Mean body-point error to the motion reference after pelvis-translation alignment\\
\bottomrule
\end{tabularx}
\end{table}

\begin{table*}[!t]
\centering
\caption{\textbf{Policy comparison in simulation.} Mean $\pm$ SD over five seeds
(100 balls and one clearance trial each). Joint: one policy; Composed:
separate footwork and striking skills; w/o arm refine.: arm refinement disabled.}
\label{tab:policy}
\footnotesize
\setlength{\tabcolsep}{2pt}
\renewcommand{\arraystretch}{1.18}
\renewcommand{\tabularxcolumn}[1]{m{#1}}
\begin{tabularx}{\textwidth}{@{}>{\centering\arraybackslash}m{0.145\textwidth}>{\centering\arraybackslash}m{0.125\textwidth}*{7}{>{\centering\arraybackslash\leavevmode}X}@{}}
\toprule
& & \multicolumn{5}{c}{\textbf{Striking}} & \textbf{Clearance} & \textbf{Motion}\\
\cmidrule(lr){3-7}\cmidrule(lr){8-8}\cmidrule(lr){9-9}
Method & Skill design
& \makecell{$SR_C$\\(\%) $\uparrow$} & \makecell{$SR_H$\\(\%) $\uparrow$} & \makecell{$E_p$\\(cm) $\downarrow$}
& \makecell{$E_v$\\(m/s) $\downarrow$} & \makecell{$E_n$\\($^\circ$) $\downarrow$}
& \makecell{$E_M$\\(cm) $\downarrow$} & \makecell{MPJPE\\(mm) $\downarrow$}\\
\midrule
\multirow{2}{*}{HITTER~\cite{su2026hitter}} & Composed & $55.20\!\pm\!{\scriptstyle 6.14}$ & $42.40\!\pm\!{\scriptstyle 8.26}$ & $12.47\!\pm\!{\scriptstyle 2.63}$ & $1.09\!\pm\!{\scriptstyle 0.14}$ & $4.83\!\pm\!{\scriptstyle 1.46}$ & $3.76\!\pm\!{\scriptstyle 1.18}$ & \na\\
 & Joint & $47.20\!\pm\!{\scriptstyle 6.76}$ & $37.80\!\pm\!{\scriptstyle 8.17}$ & $14.66\!\pm\!{\scriptstyle 3.06}$ & $1.05\!\pm\!{\scriptstyle 0.08}$ & $\mathbf{3.91}\!\pm\!{\scriptstyle 1.27}$ & $4.95\!\pm\!{\scriptstyle 1.05}$ & \na\\
\midrule
\multirow{2}{*}{SMASH~\cite{ren2026smash}} & Composed & $53.80\!\pm\!{\scriptstyle 1.92}$ & $35.80\!\pm\!{\scriptstyle 0.84}$ & $9.62\!\pm\!{\scriptstyle 0.53}$ & $1.08\!\pm\!{\scriptstyle 0.08}$ & $7.20\!\pm\!{\scriptstyle 1.28}$ & $12.54\!\pm\!{\scriptstyle 3.45}$ & $92.35\!\pm\!{\scriptstyle 1.75}$\\
 & Joint & $9.20\!\pm\!{\scriptstyle 1.79}$ & $3.80\!\pm\!{\scriptstyle 1.92}$ & $16.98\!\pm\!{\scriptstyle 0.61}$ & $1.11\!\pm\!{\scriptstyle 0.06}$ & $8.38\!\pm\!{\scriptstyle 0.45}$ & $17.75\!\pm\!{\scriptstyle 2.94}$ & $95.08\!\pm\!{\scriptstyle 2.76}$\\
\midrule
\rowcolor{black!5}
\cellcolor{white} & \textbf{Composed} & $\mathbf{67.40}\!\pm\!{\scriptstyle 4.88}$ & $\mathbf{55.60}\!\pm\!{\scriptstyle 9.76}$ & $7.43\!\pm\!{\scriptstyle 0.44}$ & $1.13\!\pm\!{\scriptstyle 0.34}$ & $8.29\!\pm\!{\scriptstyle 2.06}$ & $\mathbf{2.11}\!\pm\!{\scriptstyle 0.57}$ & $89.08\!\pm\!{\scriptstyle 0.80}$\\
 & \mbox{w/o arm refine.} & $55.20\!\pm\!{\scriptstyle 5.36}$ & $44.20\!\pm\!{\scriptstyle 12.97}$ & $7.77\!\pm\!{\scriptstyle 0.46}$ & $1.16\!\pm\!{\scriptstyle 0.32}$ & $8.72\!\pm\!{\scriptstyle 1.66}$ & $2.11\!\pm\!{\scriptstyle 0.57}$ & $\mathbf{89.04}\!\pm\!{\scriptstyle 1.72}$\\
\multirow{-3}{*}{\makecell{\textbf{PingpongDuo}\\\textbf{(ours)}}} & Joint & $59.00\!\pm\!{\scriptstyle 6.12}$ & $46.60\!\pm\!{\scriptstyle 7.50}$ & $\mathbf{6.00}\!\pm\!{\scriptstyle 1.02}$ & $\mathbf{0.62}\!\pm\!{\scriptstyle 0.16}$ & $11.74\!\pm\!{\scriptstyle 2.27}$ & $4.06\!\pm\!{\scriptstyle 3.08}$ & $93.33\!\pm\!{\scriptstyle 2.71}$\\
\bottomrule
\end{tabularx}
\end{table*}

\begin{figure*}[!t]
\centering
\includegraphics[width=\textwidth]{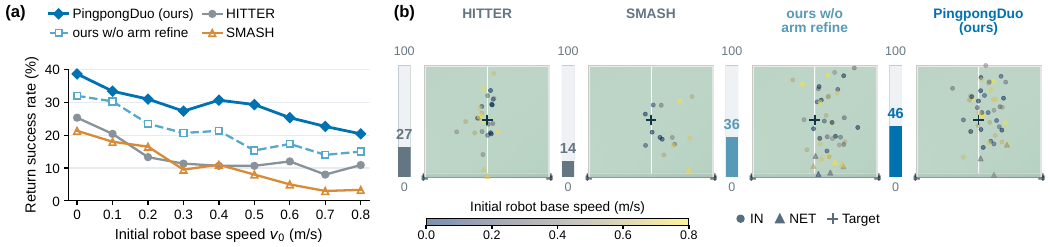}
\caption{\textbf{Arm refinement under initial motion.} (a) Return success versus
initial base speed. (b) Landings and legal-return counts from 100 trials per
method. Colors denote initial speed.}
\label{fig:motor}
\label{fig:residual}
\end{figure*}

\subsection{Experimental Setup}
We deploy two 29-DoF Unitree G1 robots with external motion capture (MoCap) for robot and
ball state estimation. Simulation uses Isaac Lab~\cite{mittal2025isaac}, with physics at 200 Hz
and control at 50 Hz. The planner benchmark uses five evaluation seeds, with
100 attempts per seed. Policy tests
use five seeds, with 100 balls and clearance trial per seed.
The refinement study uses 100 movement-and-striking trials per condition.
Results are reported as means and standard deviations across evaluation
seeds where multiple seeds are used.

\Cref{tab:metrics} summarizes the metrics. A legal return crosses the net and
lands on the opponent's table. In doubles, it must also follow the alternating
order. Racket errors are evaluated at the commanded strike time $t^\star$.
Here $n_r$ is the unit racket-face normal and
$\hat v_r^\star$ is the unit target velocity. Clearance is evaluated separately
from striking. MPJPE measures tracking of each policy's native reference.

\subsection{Team Coordination}
The full system sustains rallies with a human on hardware.
\cref{fig:multi_robot_pingpong} shows four consecutive alternating returns.
After each strike, one robot clears space while its teammate approaches
the next striking position. Clearance and preparation overlap in time.
This requires both robots to act within the same workspace without waiting
for the other to finish its entire movement. The hardware demonstrations
include sequences of more than ten consecutive alternating team strikes.

In simulation, we compare rule-based positioning, heuristic CBF filtering,
learned-risk filtering, and our learned positioning policy
in \Cref{tab:planner}. All methods share the low-level controller.
The two learned-risk variants also share the risk model and filtering
settings. Their comparison separates the effect of target selection from
the effect of filtering.

With the same learned-risk filter, learned positioning increases mean
consecutive returns from 3.79 to 4.87 compared with fixed rule-based targets. This suggests that target
selection matters beyond avoiding interference during the current shot.
The team must also retain room and time to prepare for subsequent turns.
Joint positioning accounts for both robots when choosing these targets.

\cref{fig:team} shows different coordination behavior among different planners through one single nominal example case.
The rule-based and heuristic-CBF planners fail to avoid robot contact in the shown
case. Both learned-risk variants avoid that contact while ours(RL+Learned Risk) performs the lowest strike position error, retaining best striking quality. The example highlights the trade-off between clearance and striking: the farther to move a robot away conservatively, the worse quality the striking executes, to some extent. Joint positioning RL and role-weighted filtering address these requirements together.

\subsection{Footwork and Striking Performance}
We reproduce and adapt HITTER~\cite{su2026hitter} and SMASH~\cite{ren2026smash} to doubles
movement commands. HITTER combines upper-body swing references with base
and racket task rewards. SMASH uses target-matched whole-body references.
SMASH and our method use the same reference motion library of striking and movement for fair comparison.
For each method, we compare one joint policy with separate footwork and
striking skills. Our joint or composed skill control variants use the same arm refinement policy for striking.

Across all three methods, composition improves return success in \Cref{tab:policy}. A joint policy must sustain recovery motion and then
respond to a new strike command. This may require braking or moving in the
opposite direction. Separate skills allow movement and striking objectives to be
trained individually, while transition training prepares each
skill for the body states inherited from the preceding motion.

Our composed controller combines the highest mean return rate with the
lowest clearance error. Human doubles references, including lateral shuffle
steps, guide the larger movements needed for clearance and recovery.
Compared with HITTER, PingpongDuo combines more accurate clearance with
stronger return performance. This is consistent with the use of doubles
motion references and dedicated footwork training. Our joint variant has smaller position and velocity errors but
returns fewer balls. Thus, isolated tracking metrics do not fully describe
the ability to move and return the ball in sequence.
The results highlight the importance of coordinating footwork and
striking when evaluating doubles controllers.

\subsection{Effect of Arm Refinement}
With the nominal skills fixed, arm refinement raises contact success from
55.2\% to 67.4\% and return success from 44.2\% to 55.6\%
(\Cref{tab:policy}). The improvement concerns both reaching the ball and
producing a valid return. Clearance performance is unchanged because the
refinement acts only during striking.
\cref{fig:motor}~(a) examines return success under initial motion. We set the
robot's initial motion in simulation and record its base speed $v_0$ when
the ball appears. The refined policy retains an advantage over the nominal
skills across the tested speeds. This supports the use of state-dependent
arm adjustments when footwork leaves residual body motion at strike entry.
\cref{fig:residual}~(b) complements the curves with landing distributions.
Refinement produces more legal returns in this ablation study. The landing patterns
reflect ball outcomes rather than nominal racket tracking.
This comparison evaluates whether the arm correction improves
the resulting ball return beyond tracking the planned racket command.

\section{Conclusion}
We present \methodname, a framework that combines agile
footwork and precise striking with safe team coordination
in humanoid table tennis doubles.
Joint positioning and learned risk filtering coordinate the
teammates' movements, while skill composition, transition
training, and arm refinement support accurate strikes after
repositioning.
Simulation results show improved ball-return performance
and longer alternating return sequences.
In real-world deployment, two robots perform more than 10 consecutive
alternating strikes, and complete a 2.15 m
clearance-and-recovery cycle in 2.4 s.
\par\textbf{Limitations and future work.}
Our real-world system relies on external motion capture to estimate
robot and ball states. The current sensing setup does not provide
ball-spin measurements, and spin effects are not included in the
trajectory prediction or racket--ball interaction models.
Future work will investigate onboard perception and spin
estimation. We also aim to extend the system with agents
that autonomously select both their attacking and defensive strategies
based on the evolving states of their opponents.

\section*{Acknowledgments}
We thank Yankai Liao, Junli Ren, and Penglin Fu for their insightful feedback and fruitful discussions.
We also thank Yankai Liao, Tianqi Zhang, Xinran Zhao, and the Archon Robotics team
for their help with filming the demon videos.

\bibliographystyle{IEEEtran}
\bibliography{bib_short,ref}
\end{document}